\documentclass[letterpaper, 10 pt, conference]{ieeeconf}  % Comment this line out if you need a4paper

\IEEEoverridecommandlockouts                              % This command is only needed if 
\usepackage{cite}
\usepackage{amsmath,amssymb,amsfonts}
\usepackage{algorithm}
\usepackage{algpseudocode}
\usepackage{graphicx}
\usepackage{textcomp}
\usepackage{xcolor}
\usepackage{url}
\usepackage{siunitx}
\usepackage{booktabs}
\usepackage{array}

\usepackage[caption=false]{subfig}
\usepackage{textcomp}
\usepackage{stfloats}
\usepackage{url}

\usepackage{comment}

\def\BibTeX{{\rm B\kern-.05em{\sc i\kern-.025em b}\kern-.08em
    T\kern-.1667em\lower.7ex\hbox{E}\kern-.125emX}}

\title{\LARGE \bf
Understanding User Preferences of a\\Slope-Aware Variable-Admittance Filter for a Robot Guide Dog
}
\author{Federico Esposito, Mario Selvaggio, Aaron Link, and Fabio Ruggiero% <-this % stops a space
}

\begin{document}

\maketitle
\thispagestyle{empty}
\pagestyle{empty}

\begin{abstract}
This letter investigates how the parameters of a slope-aware variable-admittance filter influence user preferences in force-based interaction with a robotic guide dog for visually impaired individuals. The proposed system consists of a quadruped robot equipped with a sensor-free rigid handle for physical guidance. The framework combines path following, momentum-based interaction-wrench estimation, and a variable-admittance filter whose stiffness and damping are adapted online from slope information extracted by the robot’s depth camera. The adaptation policies are evaluated through high-fidelity simulations and a human-subject study involving blindfolded sighted participants. Multiple strategies are compared using a Taguchi $L_9$ design of experiments. Preliminary main-effect results suggest that increasing stiffness uphill and decreasing it downhill improves both objective and subjective metrics, whereas damping shows no significant main effect.
\end{abstract}

\section{Introduction}\label{SectionIntroduction}
More than $300$ million people worldwide are blind or suffer from moderate to severe vision impairment, a figure expected to rise due to population growth and ageing~\cite{pesudovs2024global}. This condition severely affects independent mobility and quality of life.
The most common navigation aids for blind or visually impaired (BVI) people are white canes and guide dogs, both with limitations. White canes rely on physical contact, provide limited environmental feedback, and cannot detect moving or overhead obstacles. Guide dogs enable more advanced obstacle avoidance and user interaction, but are limited by high costs (about $50,000$ USD~\cite{hwang2023system}), limited availability, long training times, and reduced adaptability to unfamiliar environments~\cite{hwang2023system}.
For these reasons, robots are increasingly investigated as assistive devices for BVI individuals.
\begin{figure}[t!]
\centering
\includegraphics[width=0.8\linewidth]{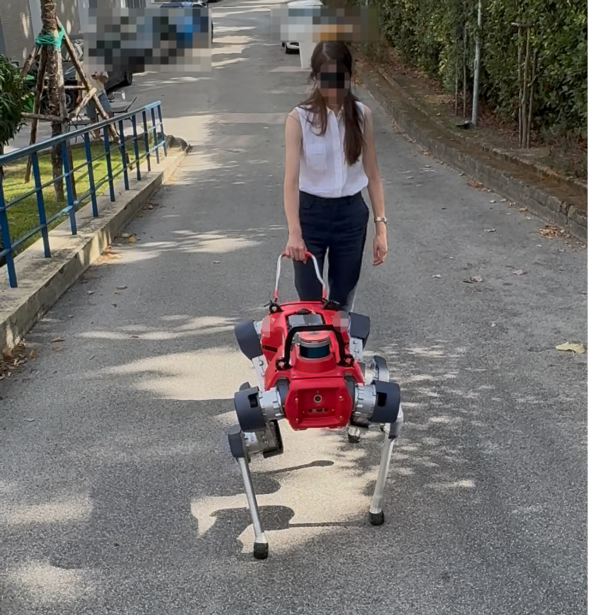}
\caption{Robot guide dog during a sloped-path human-subject trial.}
\label{fig:exp_example}
\vspace{-0.5cm}
\end{figure}
In this letter, we explore a robotic guide-dog concept for BVI users. While robots cannot replace real dogs in terms of companionship, they may support guidance through GPS navigation, speech interaction, emergency-contact features, and AI-driven perception of urban environments. This motivates sensor-minimal guidance frameworks deployable on commercial quadruped robots with limited hardware modifications. Here, we focus on the physical-interaction layer, using a rigid handle and relying only on the robot’s onboard sensing and state-estimation capabilities.

A rigid handle creates a physical connection between the quadruped robot and the user that must be safely managed.
Admittance filters are commonly used to regulate force interaction in walking aids~\cite{cho2024user} and robotic guidance systems~\cite{allenspach2022human}.
However, to the best of our knowledge, slope-aware variation of admittance parameters has only been explored in walking aids~\cite{cho2024user}, where it significantly affects user preferences in force interaction.
We extend this concept to robotic guidance, allowing the robot to adapt its compliance to different terrain conditions. Since admittance parameters vary online, both potential and kinetic energy must be considered to preserve passivity, an essential condition for stable human--robot interaction~\cite{stramigioli2015energy}.

In this letter, we consider a scenario in which a quadruped robot guides a BVI user along a predefined sloped path (see Fig.~\ref{fig:exp_example}). The human--robot two-way force interaction enables safe guidance while allowing the user to pause or deviate from the path and still receive haptic cues to return to it.
Our objective is to investigate how a terrain-slope-based variable-admittance filter influences user preferences in force-based interaction with a robotic guide dog. From basic physical intuition, walking uphill should require the robot to pull more firmly, increasing admittance stiffness and damping to stay closer to the desired path. Conversely, downhill walking should require lower parameters to avoid excessive forces that could destabilize the user. As shown later, the human-subject study supports this intuition for stiffness, while damping does not follow the same trend.

Therefore, the contributions are $(i)$ a passive slope-aware variable-admittance filter based only on the robot's onboard sensors, and $(ii)$ a human-subject study with blindfolded participants\footnote{The human-subject study has been ethically approved. Details are hidden for the double-blind process.} to fill the gap in understanding how slope-aware admittance influences user preferences in force-based interaction with a robotic guide dog.

The remainder of the letter is organized as follows. Sections~\ref{SectionState_of_the_Art}--\ref{SectionControl_Method} review related work, introduce the notation, and present the proposed framework. Sections~\ref{SectionSimulation} and~\ref{SectionHuman_Subject_Study} report simulation and experimental results, while Section~\ref{SectionDiscussion_and_Future_Directions} discusses the findings and future work.

\section{State of the Art}\label{SectionState_of_the_Art}

Adding sensors to a white cane gave birth to the idea of the smart cane, with ultrasonic and infrared sensors to extend its perception range, and a speech module to communicate the presence of obstacles to the BVI individual~\cite{hussain2016smartcane}. 
An actuated wheel was added in~\cite{slade2021multimodal} so that the cane could help the user orient themselves towards an obstacle-free direction. Recently, AI is used to provide a description of the environment to the BVI user~\cite{WEWALK}.

Another line of work concerns wearable devices providing users with information about the surroundings through vibrations or audio signals. Infrared receivers-based solutions can be used for indoor navigation~\cite{islam2018indoor}, while devices for outdoor navigation are based on ultrasonic and LIDAR sensors~\cite{bouteraa2021design}, cameras and neural networks to reconstruct the 3D environment ~\cite{bauer2020enhancing}, depth cameras~\cite{ISEEONE}, and GPS~\cite{mahendran2021computer}. Several smartphone apps have also been presented to help BVI individuals navigate indoor~\cite{fusco2020indoor} and outdoor~\cite{kuriakose2023deepnavi}. 
Speech modules often deliver information too slowly for navigation, while audio cues and vibrations can be confusing. Both methods place a high cognitive load on the user.

The idea of using wheeled robots to assist BVI individuals dates back to the late 1970s~\cite{tachi1978study}, and has recently been explored with~\cite{balatti2024robot} using an admittance-based controller. While they have several advantages over other mobile bases, like their lower energy consumption, the inability of wheels to traverse rough terrain or steps make them unsuitable for many outdoor navigation tasks.
Unmanned aerial vehicles (UAVs), by contrast, can bypass terrain constraints. Their small size and agility have prompted research into their use for guiding BVI individuals. For instance, in~\cite{allenspach2022human}, a UAV guides a blindfolded person via a tether, with path-following and admittance control algorithms ensuring safety. A similar force-based approach is used in~\cite{zhang2024aerial}, where a helium-filled balloon reduces the robot's energy consumption. 

For the task of guiding BVI individuals in navigation, legged robots offer several advantages: they are less noisy, can carry more equipment, and have longer battery life than UAVs, while retaining the ability to navigate complex environments. While some works implement low-level control~\cite{morlando2023tethering}, most rely on the robot’s built-in velocity controller and focus on higher-level strategies. 
A key distinction lies in the physical interface: some frameworks use a soft tether~\cite{defazio2023seeing}, mimicking a leash, while others, including this work, adopt a rigid connection~\cite{hwang2023system, hwang2024towards}, similar to traditional guide dog harnesses, which provides immediate haptic feedback from all robot movements. A comparison of the two approaches and a study on BVI people's preferences in terms of quadruped robots' gaits and communication protocols is presented in \cite{kim2025understanding}. 
The relevance of human-applied forces is also shown in~\cite{fan2025force}, which augments a model predictive controller with a force-based term, and in~\cite{aliotta2025understanding}, which uses a learning-based method to infer the user’s intent.

\section{Notation}\label{SectionModeling}

Let us introduce an inertial world frame, $\mathcal{F}_W=\{O_W-x_W, y_W, z_W\}$, where $O_W$ is the origin and $\{x_W, y_W, z_W\}$ are the unit axes, with $x_W,y_W$ spanning the ground plane and $z_W$ being opposite to the gravity vector.
We also define a robot's body frame, $\mathcal{F}_B=\{O_B-x_B, y_B, z_B\}$, where $O_B$ is placed at the geometrical center of the quadruped robot's torso, $x_B$ points towards the robot's heading direction, $y_B$ points towards the robot's left lateral direction, and $z_B$ completes the right-handed frame. 

In our framework, the quadruped robot is controlled via velocity commands, specifically the velocity of $\mathcal{F}_B$ with respect to $\mathcal{F}_W$, expressed in $\mathcal{F}_B$. This approach is well-suited for most commercially available quadruped robots, which typically include robust low-level controllers that map velocity inputs into coordinated leg joint movements. This design choice aligns with the vision outlined in Section~\ref{SectionIntroduction} and allows the framework to remain largely hardware-agnostic, aside from some unavoidable gain tuning.

\begin{figure}
    \centering
    \includegraphics[width=\linewidth]{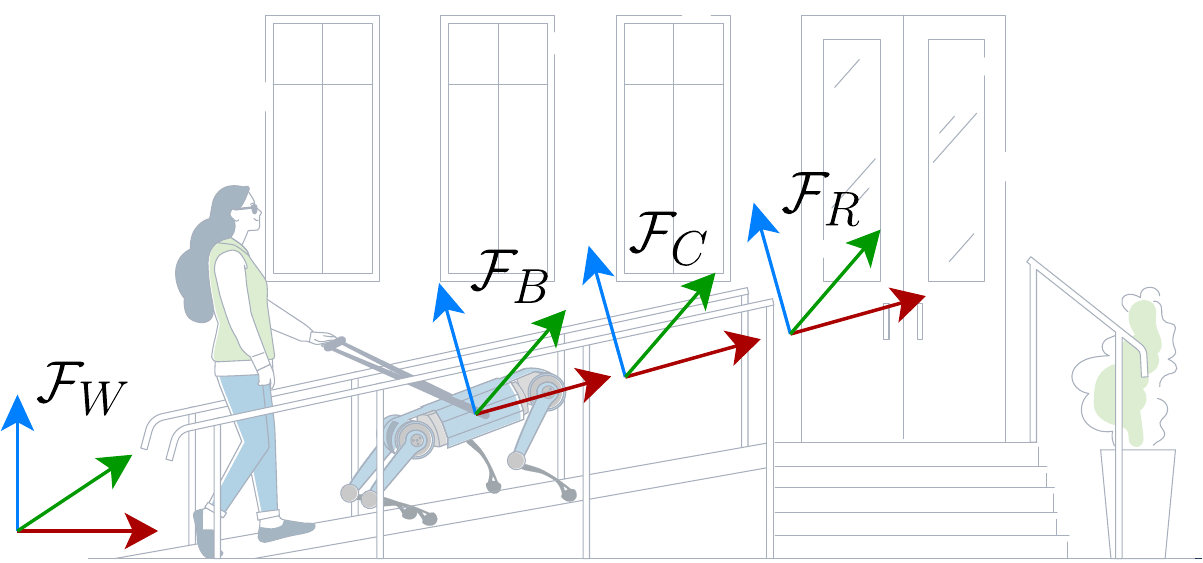}
    \caption{The four reference frames used in our framework. $\mathcal{F}_W$ is the inertial frame, $\mathcal{F}_B$ the robot's body frame, $\mathcal{F}_R$ represents the desired robot pose, and $\mathcal{F}_C$ is the compliant pose computed by the admittance filter.}
    \label{fig:Frames}
    \vspace{-0.5cm}
\end{figure}

Since an admittance filter will be used to let the robot interact with the human, it is also helpful to define two additional frames, namely, $\mathcal{F}_R=\{O_R-x_R, y_R, z_R\}$ and $\mathcal{F}_C=\{O_C-x_C, y_C, z_C\}$, which respectively represent the reference pose for the robot and the compliant pose, obtained after considering the effect of the external force on the robot. $\mathcal{F}_R$ and $\mathcal{F}_C$ are imposed so that $z_R$, $z_C$ are always parallel to $z_B$. The frames used in our framework are shown in Fig.~\ref{fig:Frames}.

Let $o_b=\begin{bmatrix}x & y & z\end{bmatrix}^T \in \mathbb{R}^3$ and $R_B \in SO(3)$ be the position of $O_B$ in $\mathcal{F}_W$ and the rotation of $\mathcal{F}_B$ with respect to $\mathcal{F}_W$, respectively.
We define $p_B \in\mathbb{R}^2\times \mathcal{S}^1$ as the pose vector containing the components $x$, $y$, and the yaw angle extracted from $R_B$. We similarly define $p_C$ and $p_R$, respectively related to the pose of $\mathcal{F}_C$ and $\mathcal{F}_R$ in $\mathcal{F}_W$. 
%We will use the dot notation for the first and second derivative of those extended position vectors.
The estimate of the terrain slope in the robot's heading direction is indicated by $\phi\in\mathbb{R}$.

\section{Framework Modules}\label{SectionControl_Method}

\subsection{Admittance filter for human-robot interaction}
In order to make the robot behave in a compliant way with respect to the human user, we introduce the following variable-admittance filter
\begin{equation}\label{eq:admittance_robot}
    M(\Ddot{p}_C-\Ddot{p}_R) + D(\phi)(\dot{p}_{C}-\dot{p}_R) + K(\phi)(p_{C}-p_R) = \hat{w},
\end{equation}
where $M,D(\phi),K(\phi)\in\mathbb{R}^{3\times3}$ are the positive definite admittance inertia, damping and stiffness matrices, respectively, $\hat{w}\in\mathbb{R}^3$ is the external wrench (a force in the first two components and a torque in the third one) at the robot's geometric center computed by the momentum-based estimator (see Section~\ref{sec:momentum-based_estimator}).
The stiffness and damping matrices are defined as functions of the terrain slope angle, allowing the robot’s behavior to adapt to environmental conditions. The specific strategy for adjusting these matrices will be explored through the human subjects study, as outlined in Section~\ref{SectionIntroduction}. 

In this letter, the variation of these parameters will be limited to the components related to the linear motion, while those related to the orientation will be kept constant. 
Therefore, $K(\phi)=\text{diag}(c_{i,k}(\phi)k_N,c_{i,k}(\phi)k_N,k_\psi)$ and $D(\phi) = \text{diag}(c_{i,d}(\phi)d_N, c_{i,d}(\phi)d_N,d_\psi)$, with \mbox{$k_N,d_N > 0$} the nominal stiffness and damping values, respectively, and \mbox{$k_\psi, d_\psi > 0$} the constant stiffness and damping terms, respectively, related to the yaw.
Each scaling factor, $c_{i,j}(\phi)>0$ with $j=\{k,d\}$, is designed based on different policies and indicated through $i = \{D, N, U\}$. In particular, we consider the nominal case in which the scaling factor is constant, $c_{N,j} = 1$, regardless of the slope. 
In the $i = U$ policy, the scaling factor is greater than $\alpha_0$ when the robot is going uphill, and is less than $\alpha_0$ otherwise,
\begin{equation}\label{eq:scaling_factor_uphill}
    c_{U,j}(\phi)=
    \begin{cases}
        \alpha_0+\alpha_1\tanh{(\alpha_2\phi)},&\phi\geq0,\;\\
        \alpha_0 + \alpha_3\tanh{(\alpha_2\phi)},&\phi<0.\; 
    \end{cases}
\end{equation}
In the $i = D$ policy, the scaling factor is greater than $\alpha_0$ when the robot is going downhill ($\phi<0$) and is less than $\alpha_0$ otherwise, 
\begin{equation}\label{eq:scaling_factor_downhill}
    c_{D,j}(\phi)=
    \begin{cases}
        \alpha_0-\alpha_3\tanh{(\alpha_2\phi)},&\phi\geq0,\;\\
        \alpha_0 - \alpha_1\tanh{(\alpha_2\phi)},&\phi<0.\; 
    \end{cases}
\end{equation}
The design of~\eqref{eq:scaling_factor_uphill} follows the intuition in Section~\ref{SectionIntroduction}, while~\eqref{eq:scaling_factor_downhill} provides the opposite behavior for comparison in the human-subject study. The $\tanh(\cdot)$ function ensures smooth saturation, and the positive $\alpha$-gains, with $\alpha_3 < \alpha_0$, shape the scaling factors.

Extracting $\dot{p}_C$ and $p_C$   from~\eqref{eq:admittance_robot}, we design velocity commands for the robot as $u=\dot{p}_C+K_{P}(p_C-p_B)$, where $K_{P}\in\mathbb{R}^{3\times3}$ is a  positive definite gain matrix. Assuming $u$ is tracked with high fidelity, yields
\begin{equation}\label{eq:vel_controller}
    \dot{p}_B \approx \dot{p}_C+K_{P}(p_C-p_B).
\end{equation}
The command $u$ is expressed in $\mathcal{F}_B$ before being used.

\subsection{Energy tank and passivity proof}\label{SubsectionPassivity_Proof}
Allowing arbitrary variations of the admittance parameters at every instant may compromise the system’s passivity and stability. In particular, changes in stiffness can inject energy into the system when $\dot{c}_{i,k}(\phi) > 0$. Inspired by~\cite{zhao2022variable}, we introduce an energy tank, which stores the energy dissipated by the system and makes it available when an increase in stiffness is required. Omitting the explicit time-dependence of the variables in the integrand, the equation that rules how the energy tank evolves is
\begin{equation}\label{eq:energy_tank}
    \begin{split}
        e(t) &=e(0)+\int_0^t\bigg[(\dot{\Bar{p}}_{C}-\dot{\Bar{p}}_R)^T\Bar{D}(\phi)(\dot{\Bar{p}}_{C}-\dot{\Bar{p}}_R)\\&-\frac{1}{2}(\Bar{p}_{C}-\Bar{p}_R)^T\dot{\Bar{K}}(\phi)(\Bar{p}_{C}-\Bar{p}_R)\bigg]\mathrm{d}\sigma,
    \end{split}
\end{equation}
where $e(t)\in\mathbb{R}$ is the energy tank's level at time $t \geq0$, the vectors $\Bar{p}_{C}, \Bar{p}_{R}\in\mathbb{R}^2$ contain the linear components of $p_{C}$ and $p_{R}$, respectively, $\Bar{D}(\phi)\in\mathbb{R}^{2\times2}$ is the matrix obtained by taking the first $2\times2$ block of $D(\phi)$, while $\dot{\Bar{K}}(\phi)\in\mathbb{R}^{2\times2}$ is the derivative of the first $2\times2$ block of $K(\phi)$.

The evolution of the energy value in~\eqref{eq:energy_tank} is governed by a set of rules that define the maximum allowable stored energy, the minimum permissible level, and the system’s behavior near this lower bound. The maximum is enforced by saturating the tank value at a predefined upper limit. To prevent the energy from falling below the minimum threshold, we first compute~\eqref{eq:energy_tank} and store the result in a temporary variable. If this value falls below the lower bound, energy-consuming operations are disallowed following
\begin{equation}\label{eq:scaling_factor_dot}
    \dot{c}_{i,k}(\phi)=
    \begin{cases}
        \displaystyle\frac{d}{dt}c_{i,k}(\phi),&e(t)\geq e_{min},\;\\
        0,&e(t)< e_{min}.\; 
    \end{cases}
\end{equation}
After this correction, the actual tank value is updated, allowing dissipative contributions to be stored. 

Choosing as storage function the following $V \geq 0$

\begin{equation}\label{eq:storage_function}
 V=\dfrac{1}{2}(\dot{p}_{C}-\dot{p}_R)^TM(\dot{p}_{C}-\dot{p}_R)+\dfrac{1}{2}(p_{C}-p_R)^TK(\phi)(p_{C}-p_R),
\end{equation}
where the first term represents the kinetic energy, and the second term the elastic potential energy of the compliant system in~\eqref{eq:admittance_robot}. 
We compute the time derivative of~\eqref{eq:storage_function} and fold~\eqref{eq:admittance_robot} and~\eqref{eq:vel_controller} into it, resulting in $\dot{V}=(\dot{p}_{B}-\dot{p}_R)^T\hat{w}-(p_C-p_B)^TK_{p}\hat{w}-(\dot{p}_{C} - \dot{p}_R)^TD(\phi)(\dot{p}_{C} - \dot{p}_R)+\frac{1}{2}({p}_{C} - {p}_R)^T\dot{K}(\phi)({p}_{C} - {p}_R).$ The passivity condition $\dot{V} \leq (\dot{p}_B-\dot{p}_R)^T\hat{w}$ is satisfied if the integral of the second, third, and fourth term in~$\dot{V}$ is non-positive across the time interval $[0, t]$. The second term disappears under the assumption that the low-level controller is able to keep a negligible error between the commanded and the measured values, as done in~\cite{tognon2021physical}. The third term is non-positive, while the sign of the fourth term is the same as the one of $\dot{c}_{i,k}(\phi)$. Exploiting~\eqref{eq:energy_tank} and~\eqref{eq:scaling_factor_dot}, we can guarantee that the controlled system described by~\eqref{eq:admittance_robot} and~\eqref{eq:vel_controller}, is passive
with storage function~\eqref{eq:storage_function},  with respect to the output-input pair $(\dot{p}_B-\dot{p}_R,\hat{w})$. 
If the assumptions of negligible tracking errors do not hold, we can only state passivity of the system~\eqref{eq:admittance_robot} with respect to $(\dot{p}_C-\dot{p}_R,\hat{w})$ using the same storage function~\eqref{eq:storage_function}.

\subsection{Path Following}\label{SubectionPath_Following}
The reference $p_R$ is generated by a path-following algorithm using the robot’s current position and desired path, avoiding time-based references that could cause excessive pulling if the user pauses. Inspired by~\cite{tognon2021physical}, we use an explicit feedforward velocity rather than a force.
 The desired path consists of linear segments and circular arcs, defined in $\mathcal{F}_W$.
For linear segments, let $s_l \in [0,1]$ be the normalized arc length. In this case, the expression of $\Bar{p}_R$ is
$\Bar{p}_{R}(s_l)= \Bar{p}_{R,l,i} + s_l(\Bar{p}_{R,l,f}-\Bar{p}_{R,l,i})$, where $\Bar{p}_{R,l,i}\in\mathbb{R}^2$ and $\Bar{p}_{R,l,f}\in\mathbb{R}^2$ are the endpoints. Along the linear segment, the desired yaw angle is constant.
For circular arcs, with normalized arc length $s_c \in [0,1]$, the reference is 
\begin{align}\label{eq:p_c}
    \Bar{p}_{R}(s_c)= \Bar{p}_{R,c} + \rho\Bigg[
    \begin{split}
        \cos{(\theta_{c,i}+s_c(\theta_{c,f}-\theta_{c,i}))} \\ 
        \sin{(\theta_{c,i}+s_c(\theta_{c,f}-\theta_{c,i}))} 
    \end{split}
    \Bigg],
\end{align}
with $\Bar{p}_{R,c} \in \mathbb{R}^2$ the center, $\rho > 0$ the radius, and \mbox{$\theta_{c,i}, \theta_{c,f} \in \mathcal{S}^1$} the start and end angles. The yaw angle keeps the robot tangential to the arc.

Once the point on the desired path with the smallest Euclidean distance from $\Bar{p}_B$ and with arc length not smaller than its previous value is found, its position and yaw define $p_R$, preventing the reference frame from moving backward along the path. Here, $\Bar{p}_B$ contains the linear components of ${p}_B$. Since the path is geometric, velocity and acceleration are chosen to match its shape. Linear segments use constant velocity and zero acceleration, while circular ones obtain velocity and acceleration by differentiating~\eqref{eq:p_c} with respect to $s_c$ and rescaling them to cap the motion speed.
In the final linear segment, velocity decreases to zero as $s_l \rightarrow 1$, with acceleration held constant until then and set to zero afterward for a gradual stop.

\subsection{Momentum-based estimator}\label{sec:momentum-based_estimator}
The estimation of the external wrench extends the momentum-based estimator from~\cite{morlando2022MED} by incorporating the linear components of the center of mass (CoM)
$
\hat{w}_{com}(t) = K_o \left( \mu(q, \dot{q}) - \int_0^t \left( h(q) + \hat{w}_{com}(\sigma) \right)\mathrm{d}\sigma \right),
$
where $K_o \in \mathbb{R}^{6 \times 6}$ is a positive definite gain matrix, $\mu(q, \dot{q}) \in \mathbb{R}^6$ is the robot’s momentum, depending on the joint positions $q \in \mathbb{R}^{n_j}$ and velocities $\dot{q} \in \mathbb{R}^{n_j}$, with $n_j > 0$ the number of joints.
The term $h(q) = -mg + J_{st,com}(q)^T f_{gr}$ depends on the robot's mass, $m > 0$, the gravity vector, $g \in \mathbb{R}^6$, the Jacobian $J_{st,com}(q) \in \mathbb{R}^{3n_{st} \times 6}$ mapping ground reaction forces to CoM wrench, and the ground reaction forces, \mbox{$f_{gr} \in \mathbb{R}^{3n_{st}}$}, where $0 < n_{st} \leq 4$ is the number of stance legs. The reconstruction of $f_{gr}$ follows~\cite{fink2020proprioceptive}.

The estimated wrench $\hat{w}$ in~\eqref{eq:admittance_robot} is constructed from the first two linear components and the last torque component of $\hat{w}_{com}(t)$, transformed into $\mathcal{F}_B$ via geometric transformations. The estimator is proven to be free from bias, although it is possible for it to confuse unmodeled terms as external forces.

\section{Simulations}\label{SectionSimulation}
The proposed framework was first validated in Gazebo's simulation environment. The robotic platform used is ANYmal D by ANYbotics. Our simulations employed the same locomotion stack as the real robot and were conducted in ANYbotics' playground environment, which features a representative mix of flat terrain and slopes. Based on the slopes, the gains in \eqref{eq:scaling_factor_uphill} and \eqref{eq:scaling_factor_downhill} have been selected as $\alpha_0=1$, $\alpha_1=1$, $\alpha_2=20$ and $\alpha_3=0.5$. The gain for the robot's velocity controller has been set to $K_P = 5I_3$, with $I_n \in \mathbb{R}^{n \times n}$ the identity matrix of proper dimensions.

In~\cite{hwang2023system}, it is shown that guide dogs exert a pulling force of around \SI{40}{N} while leading a blindfolded human user. Based on this, we selected a value of $k_N = \SI{40}{N/m}$, as it results in a pulling force equal to that measured in real animals in the steady-state case, assuming a distance of \SI{1}{m} between the origins of $\mathcal{F}_C$ and $\mathcal{F}_R$.
The matrix $M$ is defined as a diagonal matrix, with elements taken from the diagonal of the real robot’s mass matrix (extracted from the available CAD model). Specifically, the first two entries correspond to the robot’s mass $m$, while the last entry represents its moment of inertia about the $z_B$-axis.

The value of $k_\psi$ was chosen so that the ratio $k_\psi / k_N$ matches the ratio between the angular and linear terms of $M$. Finally, $d_N=2\sqrt{k_Nm}$ and $d_\psi=2\sqrt{k_\psi I_{zz}}$ were selected to ensure critical damping. 
As for the momentum-based estimator, we experimentally tuned $K_o = 5I_6$. With the chosen gains, the $-3$~dB bandwidth of the estimator is about $0.8$ Hz and the settling time at $98$\% is $0.7824$ s. 
To limit the effect of the robot's steps on the estimate, the output is passed through a discrete-time first-order low-pass filter, defined as $y_k=\alpha y_{k-1}+(1-\alpha)x_k$, with $\alpha=0.99$ and an update rate of $100$ Hz. This filter has a bandwidth of approximately $0.16$ Hz, which is lower than that of the estimator and thus dictates the overall system dynamics. Ground-truth comparisons for the estimator are present in the attached supplementary material.

The estimate of $\phi$ is obtained by applying RANSAC plane fitting to the point cloud from the robot’s front depth camera~\cite{rusu2011pcl}. For real-time execution, the cloud is downsampled by a factor of $25$, and RANSAC is applied with a $0.01$~m inlier threshold. The plane normal is extracted in $\mathcal{F}_W$, transformed into $\mathcal{F}_B$, conditioned upwards, and projected onto the sagittal plane $x_Bz_B$. The slope angle $\phi$ is then computed using atan2 on the projected components. To reduce noise and near-flat erratic estimates, a deadband sets $\phi=0$ below $0.025$ rad.

\begin{figure}
    \centering
    \includegraphics[width=\linewidth]{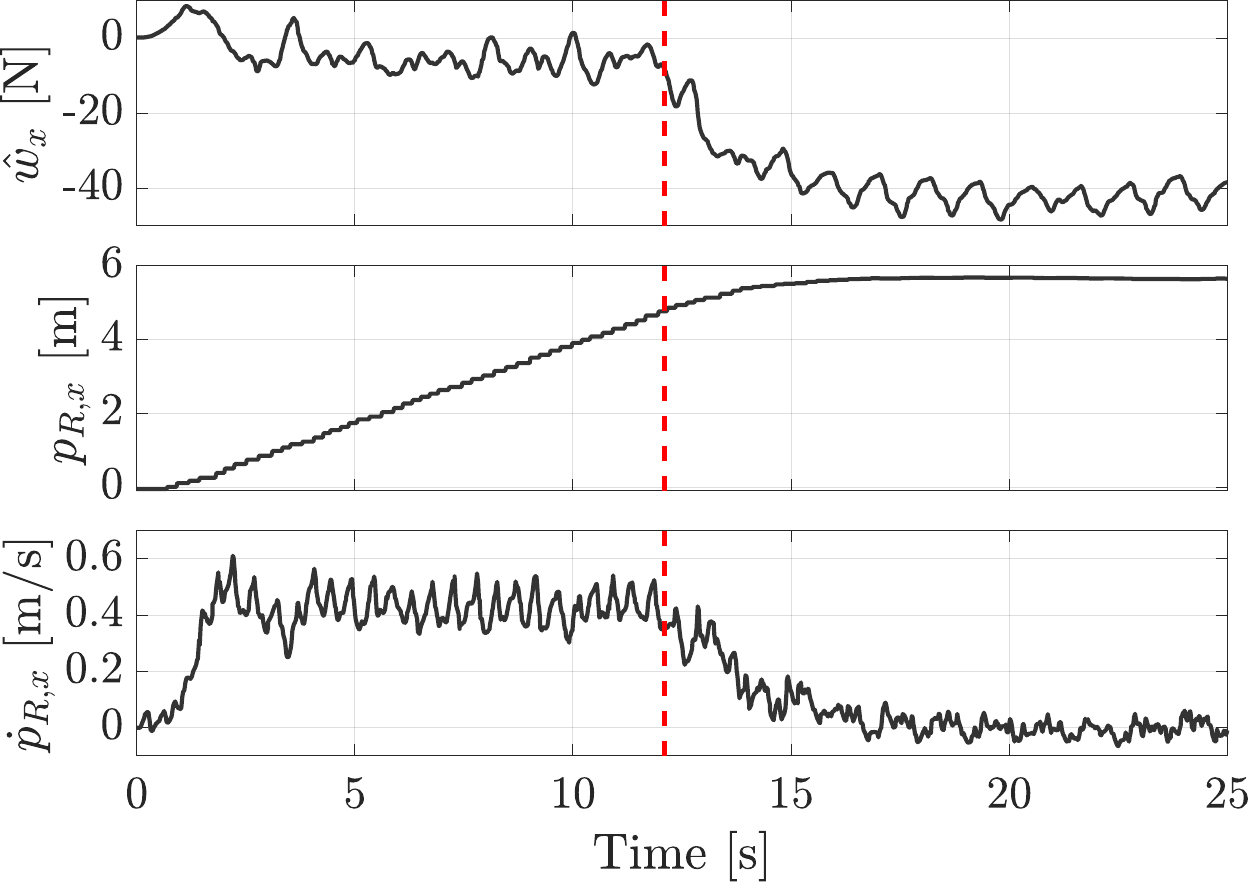}
    \caption{Top: force perceived by the wrench estimator along $x_B$. Middle: position of the robot along the $x_R$. Bottom: velocity of the robot along the $x_R$. The vertical dashed line indicates the instant when an opposing \SI{50}{N} force is applied.}
    \label{fig:Straight_plots}
    \vspace{-0.5cm}
\end{figure}

To evaluate whether the chosen parameters are reasonable, in preparation for real hardware testing, we conducted a simulation to estimate the force required to stop the robot while moving on flat terrain at a reference velocity of \SI{0.5}{m/s}, using $c_{i,k}(\phi)=c_{N,k}$ and $c_{i,d}(\phi)=c_{N,d}$. The results are shown in Fig.~\ref{fig:Straight_plots}. The vertical dashed line at $t = 12$~s indicates the moment when a force of \SI{50}{N}, opposite to motion, acts on the robot.
The top plot shows how the wrench estimator switches from measuring only frictional effects to detecting the external force. The middle and bottom plots illustrate how the robot’s position stabilizes and its velocity converges to zero. This stopping-force simulation provides a reference calibration for the selected nominal stiffness and damping values under flat-terrain conditions. Since such a force is close to the \SI{40}{N} reported in~\cite{hwang2023system}, we consider our choices for $k_N$ and $d_N$ valid.

\begin{figure}
    \centering
    \includegraphics[width=\linewidth]{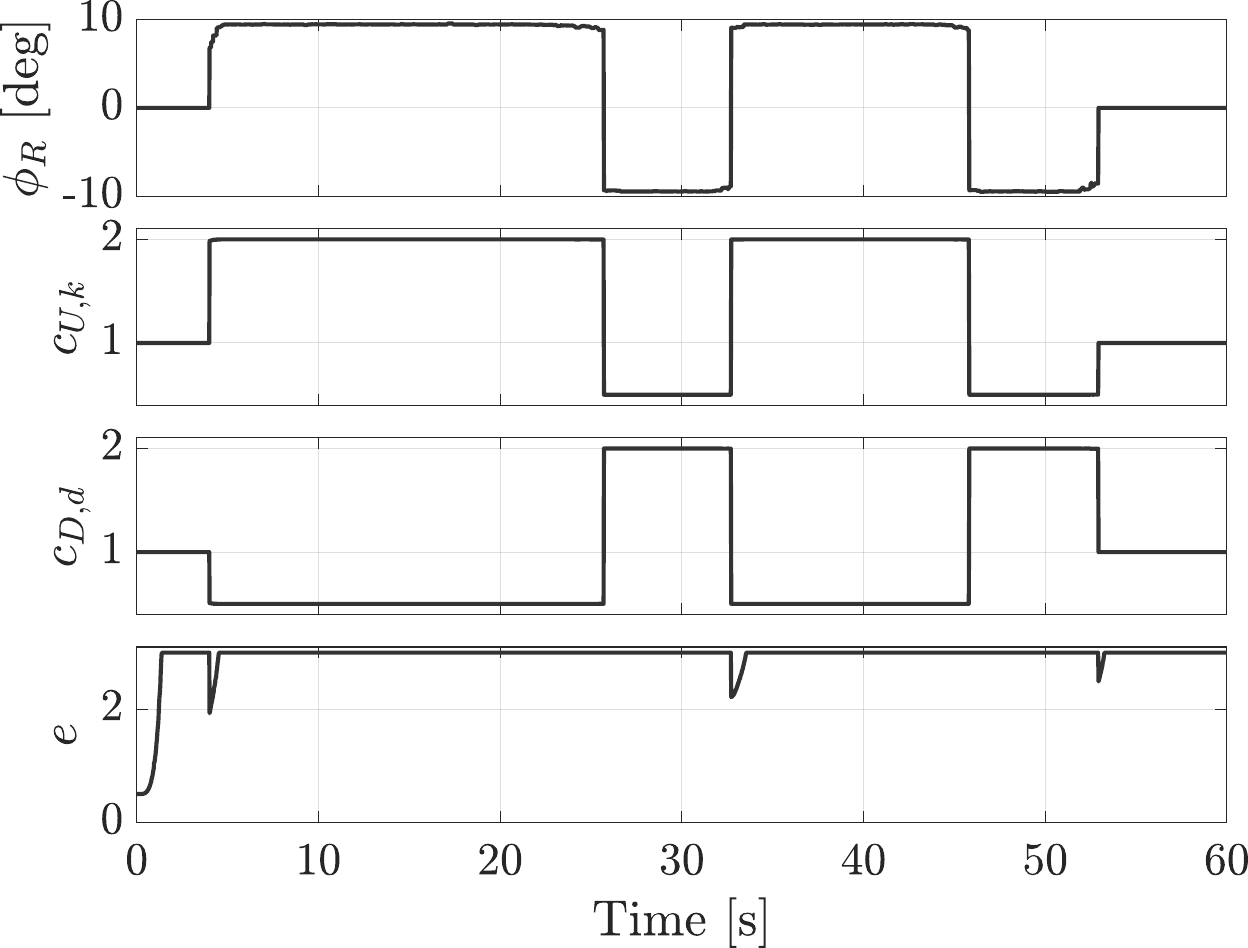}
    \caption{{First plot: slope angle $\phi$. Second plot: scaling factor used for the stiffness. Third plot: scaling factor used for the damping. Fourth plot: energy stored in the energy tank.
    }}
    \label{fig:Slope_plots}
    \vspace{-0.5cm}
\end{figure}

We performed another simulation to validate the slope estimation, the varying scaling factors, and the energy tank evolution. In Fig.~\ref{fig:Slope_plots}, we present data recorded as the robot followed a path that began and ended on flat terrain, with alternating uphill and downhill slopes. During this simulation the used scaling factors policies were $c_{U,k}(\phi)$ and $c_{D,d}(\phi)$. 
The first plot shows that $\phi$ is estimated correctly throughout the trajectory. The second and third plots display the evolution of the scaling factors for stiffness and damping, respectively. The fourth plot shows the evolution of the system’s energy, computed using~\eqref{eq:energy_tank}. The observed decreases in energy at certain time instants demonstrate the necessity of the energy tank to preserve passivity during parameter variations. The bounds for the tank \eqref{eq:energy_tank}, chosen with trial and error, are $0.1$ and $3$, while $e(0)=0.5$. 

\section{Human Subjects Study}\label{SectionHuman_Subject_Study}
This section presents a human subjects study aimed at identifying effective strategies for varying $K(\phi)$ and $D(\phi)$ in~\eqref{eq:admittance_robot}. 
A supplementary video is attached, showcasing representative trials of the experiments. 
 
\subsection{Experimental Setup Description}\label{SubsectionExperimental_Setup_Description}

Following the design philosophy outlined in Section~\ref{SectionIntroduction}, we only added a rigid handle, for the interaction with the human, to the selected ANYmal D quadruped robot. 
It is a \SI{40}{cm} long handle compatible with Julius K9 harnesses. No additional sensors were integrated into the robot: the human-robot interaction force was estimated using the momentum-based observer described earlier. We confirm the choice of the gains made in Section~\ref{SectionSimulation}.

The outdoor experiments were conducted on an asphalt terrain without irregularities that includes both approximately flat ground and a slope long $25$ m and wide $5$ m, with a maximum incline of $8.5^\circ$ (i.e., $15\%$ inclination).

\subsection{Design of Experiments}\label{SubsectionDesign_of_Experiment}

We selected three experimental factors to vary in our study, specifically 
$(i)$ \textit{Path}, the type of path the robot follows, chosen among a downhill trajectory, an uphill trajectory, and a curved path on a slope;
$(ii)$ \textit{Stiffness}, the strategy for varying $K(\phi)$ in~\eqref{eq:admittance_robot}, either increasing uphill (and thus decreasing downhill) as in~\eqref{eq:scaling_factor_uphill}, increasing downhill (and thus decreasing uphill) as in~\eqref{eq:scaling_factor_downhill}, or keeping it constant;
$(iii)$ \textit{Damping}, the strategy for varying the matrix $D(\phi)$, following the same three options as for $K(\phi)$.
Therefore, each factor has $3$ levels, which would result in $27$ possible combinations. Preparatory tests showed that a single full run takes approximately $90$ minutes, comparable to the robot’s battery life, making it impractical to test all of them. Requiring people to perform experiments for an hour and a half was deemed too stressful and fatigue-inducing for the test subjects.

To reduce the number of experiments while still gaining meaningful insights (including about untested combinations), we adopted a fractional factorial design, specifically the $L_9$ Taguchi design described in~\cite{parkrobust}. This allows us to test only $9$ combinations while preserving balanced coverage across the factor levels, but limits our results to the factors' main effects.
We define the three \textit{Path} types as $A_0$ (downhill), $A_1$ (curve), and $A_2$ (uphill). Similarly, \textit{Stiffness} variation strategies are denoted as $B_0$ (increasing downhill and decreasing uphill), $B_1$ (constant stiffness), and $B_2$ (increasing uphill and decreasing downhill). \textit{Damping} variation strategies are labeled $C_0$, $C_1$, and $C_2$ in the same manner. In order to achieve the required balance in our test cases, we selected the tests described in Table~\ref{tab:description_experiments}. 
The same three paths were repeated exactly for all subjects.

\begin{table}[t]
\centering
\caption{Description of the Experiments}
\label{tab:description_experiments}
\vspace{-3mm}
\begin{tabular}{l
  >{\centering\arraybackslash}m{0.10\linewidth}
  >{\centering\arraybackslash}m{0.12\linewidth}
  >{\centering\arraybackslash}m{0.18\linewidth}
  >{\centering\arraybackslash}m{0.18\linewidth}}
\toprule
 & Levels & \textit{Path} $A$ & \textit{Stiffness} $B$ & \textit{Damping} $C$ \\
\midrule
Test 1 & $A_0B_0C_0$ & downhill & eq. \eqref{eq:scaling_factor_downhill} & eq. \eqref{eq:scaling_factor_downhill} \\
\midrule
Test 2 & $A_0B_1C_1$ & downhill & constant & constant \\
\midrule
Test 3 & $A_0B_2C_2$ & downhill & eq. \eqref{eq:scaling_factor_uphill} & eq. \eqref{eq:scaling_factor_uphill} \\
\midrule
Test 4 & $A_1B_1C_0$ & curve & constant & eq. \eqref{eq:scaling_factor_downhill} \\
\midrule
Test 5 & $A_1B_2C_1$ & curve & eq. \eqref{eq:scaling_factor_uphill} & constant \\
\midrule
Test 6 & $A_1B_0C_2$ & curve & eq. \eqref{eq:scaling_factor_downhill} & eq. \eqref{eq:scaling_factor_uphill} \\
\midrule
Test 7 & $A_2B_2C_0$ & uphill & eq. \eqref{eq:scaling_factor_uphill} & eq. \eqref{eq:scaling_factor_downhill} \\
\midrule
Test 8 & $A_2B_0C_1$ & uphill & eq. \eqref{eq:scaling_factor_downhill} & constant \\
\midrule
Test 9 & $A_2B_1C_2$ & uphill & constant & eq. \eqref{eq:scaling_factor_uphill} \\
\bottomrule
\end{tabular}
\vspace{-0.3cm}
\end{table}

\subsection{Policy of the Human Subjects Study and Metrics}\label{SubsectionExperimental_Policy}
To reduce subjective biases such as personal preferences and ordering effects, we conducted the $9$ selected experimental cases with $15$
human subjects ($12$ men and $3$ women), with a mean age of $29.07$ years and a standard deviation of $3.88$. 
Given the exploratory nature of this study and the use of a Taguchi $L_9$ design to evaluate main effects, an a priori power analysis was conducted to determine the appropriate sample size. To achieve $80$\% power at a $0.05$ significance level to detect a large effect size (Cohen’s $f = 0.4$) across three factor levels, an initial sample size of $12$ participants was calculated assuming a parametric test, i.e., a repeated measures ANOVA. To account for the reduced statistical power inherent in non-parametric rank-based methods, this target was conservatively increased by $25$\%, justifying our final sample size of $15$ participants.
The subjects were familiar with robotics, but not with our framework, and the test order was randomized for each participant. They had no severe visual impairment or blindness and were blindfolded to simulate such conditions. Before participation, each subject signed an informed consent form covering both the experiment and personal-data handling, including video recordings. The procedure was explained in advance, including the different paths and the instruction that they would be asked to stop during the walk.

Each session began with a $5$-minute familiarization period, during which the subject interacted with the robot without the blindfold under the experimenter’s supervision. The subject could test a parameter combination excluded from the main study and walk along short uphill, downhill, and curved segments.
Once comfortable, the formal experiment started. For each test, the subject was guided to the start, blindfolded, and asked to grasp the robot’s handle. The test then followed the assigned randomized order. About $20$ seconds into each trial, the experimenter instructed the subject to stop, wait $3$ seconds, and resume walking, to help perceive the two-way force interaction enabled by our framework.

Each subject performed two tests consecutively, after which they completed a questionnaire whose questions are:
\begin{itemize}
    \item How much control did you feel you had over the robot, on a scale from 1 (very little) to 5 (a lot)?
    \item How much mental fatigue did you feel while guided by the robot, on a scale from 1 (very little) to 5 (a lot)? 
    \item How safe did you feel while guided by the robot, on a scale from 1 (very unsafe) to 5 (very safe)? 
    \item How aggressive did the robot seem, on a scale from 1 (not at all) to 5 (very aggressive)? 
    \item How much frustration did you experience while using the robot, on a scale from 1 (very little) to 5 (a lot)?
\end{itemize}

From each of these questions we extracted a metric to evaluate our framework during the experiments. They are Control ($\mathcal{C}$), Fatigue ($\mathcal{F}$), Safety ($\mathcal{S}$), Aggressiveness ($\mathcal{A}$), and Frustration ($\mathcal{R}$), respectively.
As a sixth metric, we introduced a quantitative measure to capture how much the user pushed or pulled on the robot's handle during the trial. We refer to this metric as Interaction ($\mathcal{I}$), defined as the mean across the duration of the test of the norm of the linear components of the estimated interaction wrench $\hat{w}$.

\subsection{Results}\label{SubsectionResults}
Descriptive statistics for the test results under all conditions are available in the attached supplementary material. Normality was assessed using the Shapiro--Wilk test, indicating that non-parametric tests were appropriate. Due to the Taguchi design, only main effects can be studied; therefore, we used Friedman tests with Nemenyi post-hoc tests and Tukey--Kramer correction, averaging user-level data over one factor at a time. For example, to assess \textit{Path}, we averaged Tests $1$--$3$ for downhill path $A_0$, Tests $4$--$6$ for curve path $A_1$, and Tests $7$--$9$ for uphill path $A_2$. Tables~\ref{tab:Control_friedman}--\ref{tab:Interaction_friedman} report the Friedman analyses, including Kendall's W effect sizes. A factor is considered significant for $p<0.05$.

Based on the results in Tables~\ref{tab:Control_friedman}-\ref{tab:Interaction_friedman}, the following can be concluded. The \textit{Path} and \textit{Stiffness} factors are significant in $\mathcal{C}$. For the $\mathcal{F}$ metric, no factor is significant.
Only \textit{Path} is significant in $\mathcal{S}$. For metric $\mathcal{A}$, only \textit{Stiffness} is significant. No factor is significant for metric $\mathcal{R}$. Finally, both \textit{Path} and \textit{Stiffness} are significant for metric $\mathcal{I}$.

\begin{figure}[t!]
\centering
\subfloat[]{
    \includegraphics[width=0.32\linewidth, trim={0 0 5px 12px},clip]{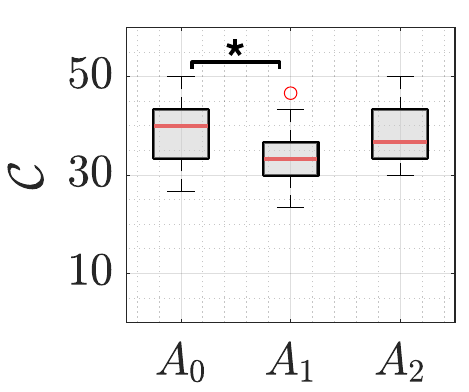}
}
\subfloat[]{
    \includegraphics[width=0.32\linewidth, trim={0 0 5px 12px},clip]{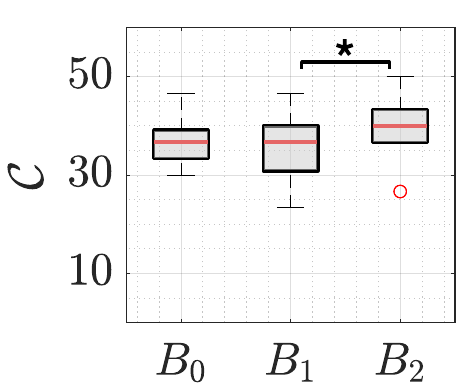}
}
\subfloat[]{
    \includegraphics[width=0.32\linewidth, trim={0 0 5px 12px},clip]{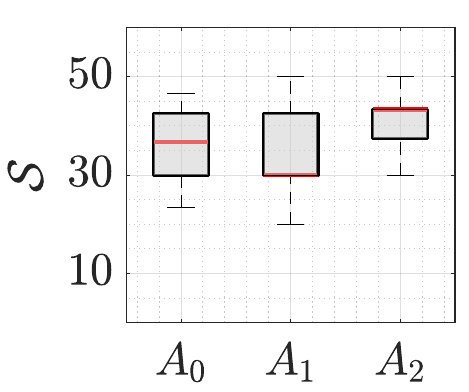}
}

\subfloat[]{
    \includegraphics[width=0.32\linewidth, trim={0 0 5px 12px},clip]{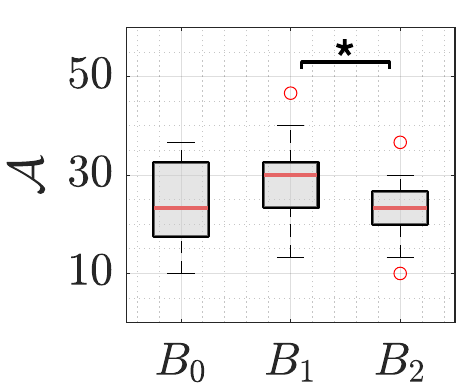}
}
\subfloat[]{
    \includegraphics[width=0.32\linewidth, trim={0 0 5px 12px},clip]{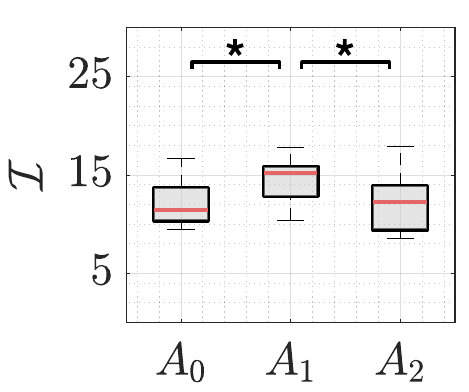}
}
\subfloat[]{
    \includegraphics[width=0.32\linewidth, trim={0 0 5px 12px},clip]{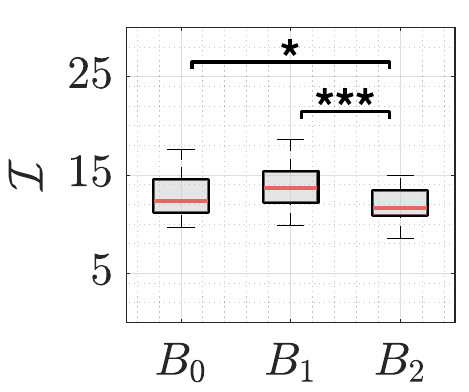}
}

\caption{Boxplots detailing the distribution of metrics across significant factors. The central red lines indicates the medians, and the bottom and top edges of the boxes represent the 25th and 75th percentiles, respectively. Whiskers extend to the most extreme inliers, while outliers are plotted as red circles. Asterisks denote statistically significant differences between factor levels based on post-hoc Nemenyi tests (* $p < 0.05;$ *** $p < 0.0005$). (a) Control across Path levels. (b) Control across Stiffness levels. (c) Safety across Path levels. (d) Aggressiveness across Stiffness levels. (e) Interaction across Path levels. (f) Interaction scores across Stiffness levels.}\label{fig:Metrics_bars}
\vspace{-0.5cm}
\end{figure}

Performing Nemenyi post-hoc test on the statistically significant factor--metric pairs, we can further characterize the data we gathered. With respect to Control ($\mathcal{C}$), the curved path ($A_1$) is worse than the downhill path ($A_0$) with a $p$-value of $0.0114$, while increasing \textit{Stiffness} uphill ($B_2$) is better than keeping it constant ($B_1$) with a $p$-value of $0.0335$. For Safety ($\mathcal{S}$), the most significant comparison shows that the uphill path ($A_2$) is safer than the curved one ($A_1$), but the $p$-value of $0.0545$ is slightly above the selected threshold. Keeping \textit{Stiffness} constant ($B_1$) rather than increasing it uphill ($B_2$) increases Aggressiveness ($\mathcal{A}$) with a $p$-value of $0.0076$. With respect to Interaction ($\mathcal{I}$), the average interaction force is higher on the curve path ($A_1$) than it is on the downhill ($A_0$) and the uphill ($A_2$) one, with respective $p$-values $0.0464$ and $0.0285$, while it is smaller with \textit{Stiffness} increasing uphill ($B_2$) than it is with \textit{Stiffness} increasing downhill ($B_0$) or remaining constant ($B_1$), with respective $p$-values $0.0170$ and $0.0004$. Relative plots can be found in Fig.~\ref{fig:Metrics_bars}.

\begin{table}[t]
\centering
\caption{Friedman tests on Control}
\vspace{-3mm}\begin{tabular}{ccccccc}\toprule\label{tab:Control_friedman}
Source & Sum Sq. & d.f. &  Mean Sq. & Chi-sq & p-value & W\\
\midrule
Path & 7.2333 & \phantom{0}2  & 3.6167 & 8.5098 & 0.0142 & 0.2837\\
Error & 18.2667 & 28 & 0.6524  &  &  \\
Total & 25.5000 & 44 &  &   &  \\
\midrule
Stiffness & 5.4333 & \phantom{0}2 & 2.7167 & 6.3922 & 0.0409 & 0.2131\\
Error & 20.0667 & 28 & 0.7167 &  &  \\
Total & 25.5000 & 44 &  &  &  \\
\midrule
Damping & 3.2333 & \phantom{0}2 & 1.6167 & 3.5273 & 0.1714 & 0.1176\\
Error & 24.2667 & 28 & 0.8667 &  &  \\
Total & 27.5000 & 44 &  &  &  \\
\bottomrule
\end{tabular}
\vspace{-0.223cm}
\end{table}

\begin{table}[t]
\centering
\caption{Friedman tests on Fatigue}
\vspace{-3mm}\begin{tabular}{ccccccc}\toprule\label{tab:Fatigue_friedman}
Source & Sum Sq. & d.f. & Mean Sq. & Chi-sq & p-value & W\\
\midrule
Path & 1.4333 & \phantom{0}2 & 0.7167 & 2.7742 & 0.2498 & 0.0925\\
Error & 14.0667 & 28 & 0.5024 &  &  \\
Total & 15.5000 & 44 &  &  &  \\
\midrule
Stiffness & 1.6333 & \phantom{0}2 & 0.8167 & 4.2609 & 0.1188 & 0.1420\\
Error & 9.8667 & 28 & 0.3524 &  &  \\
Total & 11.5000 & 44 &  &  &  \\
\midrule
Damping & 1.2333 & \phantom{0}2 & 0.6167 & 2.9600 & 0.2276 & 0.0987\\
Error & 11.2667 & 28 & 0.4024 &  &  \\
Total & 12.5000 & 44 &  &  &  \\
\bottomrule
\end{tabular}
\vspace{-0.3cm}
\end{table}

\begin{table}[t]
\centering
\caption{Friedman tests on Safety}
\vspace{-3mm}\begin{tabular}{ccccccc}\toprule\label{tab:Safety_friedman}
Source & Sum Sq. & d.f. & Mean Sq. & Chi-sq & p-value & W\\
\midrule
Path & 5.7000 & \phantom{0}2 & 2.8500 & 6.3333 & 0.0421 & 0.2111\\
Error & 21.3000 & 28 & 0.7607 &  &  \\
Total & 27.0000 & 44 &  &  &  \\
\midrule
Stiffness & 4.1333 & \phantom{0}2 & 2.0667 & 5.3913 & 0.0675 & 0.1797\\
Error & 18.8667 & 28 & 0.6738 &  &  \\
Total & 23.0000 & 44 &  &  &  \\
\midrule
Damping & 1.7333 & \phantom{0}2 & 0.8667 & 2.3636 & 0.3067 & 0.0788\\
Error & 20.2667 & 28 & 0.7238 &  &  \\
Total & 22.0000 & 44 &  &  &  \\
\bottomrule
\end{tabular}
\vspace{-0.3cm}
\end{table}

\begin{table}[t]
\centering
\caption{Friedman tests on Aggressiveness}
\vspace{-3mm}\begin{tabular}{ccccccc}\toprule\label{tab:Aggressiveness_friedman}
Source & Sum Sq. & d.f. & Mean Sq. & Chi-sq & p-value & W\\
\midrule
Path & 0.7000 & \phantom{0}2 & 0.3500 & 0.7368 & 0.6918 & 0.0246\\
Error & 27.8000 & 28 & 0.9929 &  &  \\
Total & 28.5000 & 44 &  &  &  \\
\midrule
Stiffness & 7.6000 & \phantom{0}2 & 3.8000 & 9.1200 & 0.0105 & 0.3040\\
Error & 17.4000 & 28 & 0.6214 &  &  \\
Total & 25.0000 & 44 &  &  &  \\
\midrule
Damping & 1.2000 & \phantom{0}2 & 0.6000 & 1.3091 & 0.5197 & 0.0436\\
Error & 26.3000 & 28 & 0.9393 &  &  \\
Total & 27.5000 & 44 &  &  &  \\
\bottomrule
\end{tabular}
\vspace{-0.3cm}
\end{table}

\begin{table}[t]
\centering
\caption{Friedman tests on Frustration}
\vspace{-3mm}\begin{tabular}{ccccccc}\toprule\label{tab:Frustration_friedman}
Source & Sum Sq. & d.f. & Mean Sq. & Chi-sq & p-value & W\\
\midrule
Path & 1.3000 & \phantom{0}2 & 0.6500 & 2.0526 & 0.3583 & 0.0684\\
Error & 17.7000 & 28 & 0.6321 &  &  \\
Total & 19.0000 & 44 &  &  &  \\
\midrule
Stiffness & 0.0333 & \phantom{0}2 & 0.0167 & 0.0541 & 0.9733 & 0.0018\\
Error & 18.4667 & 28 & 0.6595 &  &  \\
Total & 18.5000 & 44 &  &  &  \\
\midrule
Damping & 0.0333 & \phantom{0}2 & 0.0167 & 0.0513 & 0.9747 & 0.0017\\
Error & 19.4667 & 28 & 0.6952 &  &  \\
Total & 19.5000 & 44 &  &  &  \\
\bottomrule
\end{tabular}
\vspace{-0.3cm}
\end{table}

\begin{table}[t]
\centering
\caption{Friedman tests on Interaction}
\vspace{-3mm}\begin{tabular}{ccccccc}\toprule\label{tab:Interaction_friedman}
Source & Sum Sq. & d.f. & Mean Sq. & Chi-sq & p-value & W\\
\midrule
Path & 8.1333 & \phantom{0}2 & 4.0667 & 8.1333 & 0.0171 & 0.2711\\
Error & 21.8667 & 28 & 0.7810 &  &  \\
Total & 30.0000 & 44 &  &  &  \\
\midrule
Stiffness & 15.6000 & \phantom{0}2 & 7.8000 & 15.6000 & 0.0004 & 0.5200\\
Error & 14.4000 & 28 & 0.5143 &  &  \\
Total & 30.0000 & 44 &  &  &  \\
\midrule
Damping & 0.1333 & \phantom{0}2 & 0.0667 & 0.1333 & 0.9355 & 0.0044\\
Error & 29.8667 & 28 & 1.0667 &  &  \\
Total & 30.0000 & 44 &  &  &  \\
\bottomrule
\end{tabular}
\vspace{-0.3cm}
\end{table}

\section{Discussion and Future Directions}\label{SectionDiscussion_and_Future_Directions}

After objectively presenting the results, we now offer our conclusions and outline directions for future research.

Tables~\ref{tab:Fatigue_friedman} and~\ref{tab:Frustration_friedman} indicate that no factor has a statistically significant influence on $\mathcal{F}$ or $\mathcal{R}$. Combined with the fact that both metrics consistently report average values below $2$, we conclude that our framework does not induce stress or mental fatigue, independent of the chosen control parameters.

The \textit{Path} significantly influences multiple metrics, including $\mathcal{C}$, $\mathcal{S}$, and $\mathcal{I}$. In particular, the curved path consistently performed worse across several of them. This suggests that more attention should be given to the robot’s behavior outside linear segments. Future work could explore extending the variable-admittance filter to angular components or investigating how different nominal values of angular inertia, stiffness, and damping affect the user.

The fact that increasing the stiffness uphill, as in equation \eqref{eq:scaling_factor_uphill}, has positive effects on Control ($\mathcal{C}$), Aggressiveness ($\mathcal{A}$), and Interaction ($\mathcal{I}$) supports our initial intuition: in a robotic guide dog context, varying the admittance stiffness yields better results than keeping it fixed. Increasing stiffness uphill helps the robot stay closer to the reference path and better guide the user, while lower stiffness downhill reduces the risk of pulling and potential imbalance. Lower values of Interaction are seen as positive because with them the robot can perform its guidance task while giving less physical stimulation to the human user.

Contrary to the findings of~\cite{cho2024user}, varying the damping did not yield significant effects. This discrepancy can be attributed because our system explicitly guides the user along a predefined path, with stiffness playing a more critical role. We find instead coherence between our findings and those of \cite{kim2025understanding}, since our study supports that the role of compliance (in our case mainly the stiffness) significantly impacts users' preferences in terms of interaction with a robot guide dog.

The data in Section~\ref{SubsectionResults} also offer insight into future improvements. Since Control ($\mathcal{C}$), Safety ($\mathcal{S}$) and Interaction ($\mathcal{I}$) were strongly influenced by the \textit{Path}, future evaluations should use a consistent path, ideally combining uphill and downhill segments, to better isolate effects related to the controller's parameters from path geometry.

The \textit{Stiffness} factor was found to be significant for Control ($\mathcal{S}$), Aggressiveness ($\mathcal{A}$), and Interaction ($\mathcal{I}$). Since all three metrics favor stiffness increasing uphill, a broader range of stiffness values should be tested to better capture its effects and to fine-tune how it should evolve with respect to the slope angle. Additionally, modifying the nominal stiffness $k_N$ (which remained constant in this study) could provide valuable insights, though this was previously avoided to limit the total number of experiments.

Within the tested range and the preliminary main-effect analysis enabled by the $L_9$ design, damping showed no statistically significant effect, suggesting that future studies may prioritize stiffness. Future work will also study stiffness--damping interaction and coupling effects without relying on a Taguchi design.

The discrepancy between our a priori sample size assumptions and the fact that most of the significant effect sizes our study found are small to medium (Kendall's W between $0.1$ and $0.5$), however, tells us that future studies will require more participants to  capture more nuanced dynamics or to definitively confirm the negligible impact of damping.

The proposed framework could be integrated into a broader navigation architecture including obstacle avoidance,
route selection, user-intent handling, hazard detection, and safety supervision. In such an architecture, high-level
modules would generate the navigation plan, which would then be provided to the variable-admittance module
as the reference path $p_R$. The parameter-adaptation policy could also exploit information from these high-level
modules, for instance by increasing stiffness near obstacles or hazards.

Finally, as blindfolded sighted users may differ from BVI users in mobility strategy, trust calibration, tactile interpretation, balance behavior, and prior experience with canes or guide dogs, future studies should involve actual BVI participants to reduce the potential bias introduced by simulated visual impairment and to assess whether the observed trends generalize to the intended user population.

\bibliographystyle{IEEEtran}
\bibliography{biblio}

\end{document}